\documentclass[draftcls,onecolumn]{IEEEtran}
\ifCLASSINFOpdf
  \usepackage[pdftex]{graphicx}
\else
  \usepackage[dvips]{graphicx}
\fi
\usepackage{amsmath}
\newcommand{\abs}[1]{\left\lvert #1 \right\rvert}
\ifCLASSOPTIONcompsoc
  \usepackage[caption=false,font=normalsize,labelfont=sf,textfont=sf]{subfig}
\else
  \usepackage[caption=false,font=footnotesize]{subfig}
\fi

\usepackage{multirow}
\usepackage{hyperref}
\usepackage{makecell}

\begin{document}
%
\title{Low-Thrust Orbital Transfer using Dynamics-Agnostic Reinforcement Learning}
%
%
%

\author{Carlos~M.~Casas,~
        Belén~Carro,~
        and~Antonio~Sánchez-Esguevillas
\thanks{Carlos M. Casas, Belén Carro and Antonio Sánchez-Esguevillas are with the Department TSyCeIT, ETSIT, University of Valladolid, Spain.}
\thanks{© 20xx IEEE. Personal use of this material is permitted. Permission from IEEE must be obtained for all other uses, in any current or future media, including reprinting/republishing this material for advertising or promotional purposes, creating new collective works, for resale or redistribution to servers or lists, or reuse of any copyrighted component of this work in other works.}
}

%
%

\markboth{IEEE}%
{Shell \MakeLowercase{\textit{et al.}}: Low-Thrust Orbital Transfer using Dynamics-Agnostic Reinforcement Learning}
%



\maketitle

\begin{abstract}
Low-thrust trajectory design and in-flight control remain two of the most challenging topics for new-generation satellite operations. Most of the solutions currently implemented are based on reference trajectories and lead to sub-optimal fuel usage. Other solutions are based on simple guidance laws that need to be updated periodically, increasing the cost of operations. Whereas some optimization strategies leverage \emph{Artificial Intelligence} methods, all of the approaches studied so far need either previously generated data or a strong \textit{a priori} knowledge of the satellite dynamics. This study uses model-free \emph{Reinforcement Learning} to train an agent on a \emph{constrained} pericenter raising scenario for a low-thrust medium-Earth-orbit satellite. The agent does not have any prior knowledge of the environment dynamics, which makes it unbiased from classical trajectory optimization patterns. The trained agent is then used to design a trajectory and to autonomously control the satellite during the cruise. Simulations show that a dynamics-agnostic agent is able to learn a quasi-optimal guidance law and responds well to uncertainties in the environment dynamics. The results obtained open the door to the usage of \emph{Reinforcement Learning} on more complex scenarios, multi-satellite problems, or to explore trajectories in environments where a reference solution is not known.
\end{abstract}


%
\IEEEpeerreviewmaketitle

\section{Introduction}
%
%
%
%

\IEEEPARstart{S}{atellites} require a large initial investment to build and place in the desired orbit. For that reason, a successful mission is usually exploited way beyond the initial expected operational life. One of the main factors that restricts the lifetime of a satellite is the available fuel on-board, mainly used to change or correct its trajectory. Therefore, optimising the fuel consumption is of the utmost importance.

Traditionally, satellites are equipped with thrusters that produce a large force during a short amount of time, in the order of seconds, sometimes minutes. Control problems typically need to find out the optimal direction of the thrust and the start and end of the manoeuvre. Since the duration is expected to be short, the direction of the thrust is either constant or follows a simple parametric guidance law. Due to the low number of free variables and the well-known dynamics throughout the environment, these problems are relatively simple to solve, either mathematically or numerically. An experienced flight dynamics engineer can develop some sort of intuition of what a good initial guess for the parameters can be, so the optimization problems only require a local search around that initial guess.

Some satellites already equip low-thrust devices, such as electric propulsion systems or solar sails, and their use is being generalised. Scientific missions have used them for a few years, the latest examples being \emph{ESA/JAXA}'s \emph{BepiColombo} \cite{garcia2006bepi} launched on 2018 currently on its way to Mercury; and \emph{NASA}'s \emph{NEA Scout} \cite{mcnutt2014neascout} scheduled for launch in 2021. In the commercial sector, some companies like \emph{OHB System} \cite{lubberstedt2014electric} or \emph{Boeing} \cite{feuerborn2013finding} have started to build satellite platforms fully based on electric propulsion for geostationary satellites. Mega-constellations of small satellites on low-Earth orbit such as \emph{SpaceX}'s \emph{Starlink} \cite{potrivitu2020review} also have a similar architecture.

Low-thrust devices produce a small force that can be maintained for a long period of time, even days or weeks. On the one hand, they end up spending much less fuel than a traditional propulsion system for the same impulse, thus increasing the lifetime or range of the satellite. On the other hand, the control problems are much more difficult to solve, being numerical methods virtually the only way to achieve an acceptable solution. Moreover, it is difficult for a flight dynamics engineer to give an initial guess on what the optimal solution will be, so a thorough exploration phase needs to be performed. Thus the need to find a brand new set of tools to find near-optimal solutions.

The most commonly used approach for optimizing low-thrust trajectories consists on dividing the trajectory into arcs and optimising them separately (e.g. \cite{novak2011design, whiffen2006mystic}), each of them with its own parametric law. If not enough laws are defined then the trajectory is sub-optimal; and if too many laws are defined then the number of free variables is very large, the computation time increases massively, and the problem becomes intractable. Some methods have been developed to solve these problems, such as \emph{Discrete Mechanics and Optimal Control} (\emph{DMOC}) \cite{junge2005dmoc} where the original problem is transformed into a linear constrained optimization problem with thousands of free variables representing position and force on intermediate points, and constraints represent the approximate dynamics of the problem; or \emph{Evolutionary Algorithms} such as genetic algorithms or swarm optimization \cite{shirazi2017evolutionary}, where potential solutions are coded and mixed together to produce new candidate solutions.

All these methods heavily rely on the knowledge of the dynamics and on the definition of astute intermediate checkpoints. On top of that, the obtained solutions do not adapt to changes in the problem conditions. The control problem needs to be re-solved every few days after the latest observations of the satellite have been received, so that the free parameters are re-tuned. In order to improve the autonomy of satellites, trajectory control has been looking into the \emph{Artificial Intelligence} field \cite{izzo2019ml}. Some attempts consist on approximating certain aspects of the trajectory, such as the cost function or the guidance law, using \emph{Supervised Learning} \cite{perez2021ml}; or training an agent based on pre-created failure scenarios \cite{rubinsztejn2019nn}. This requires a lot of ad-hoc created data, and the outcome is a trajectory that just replicates what it has been fed with. Other studies used several aspects of \emph{Artificial Intelligence} to explore the space of possible arcs and design a trajectory to be refined by other means \cite{dasstuart2018rapid}. Finally, a few studies tried to directly train an agent using \emph{Reinforcement Learning} \cite{miller2019rl}, but they heavily relied on the knowledge of the problem dynamics, up to the point to reward the agent at each step depending on the distance to a target trajectory.

The goal of this article is to study the feasibility of a simpler more direct \emph{Reinforcement Learning} approach, where an agent is left free to act the thrusters of the satellite in a particular environment without intermediate rewards. The agent does not have any \emph{a priori} knowledge of the dynamics and it does not make any assumptions about the best coordinate system to represent the problem, making it totally unbiased from any classical control pattern. After being trained, the agent is used to create the full quasi-optimal trajectory, and to control the satellite all along the trajectory without any further update on the agent parameters, resulting in a totally autonomous fault-resilient satellite.

Note that the novelty of this study lies on the simplicity of the method used to obtain an optimal guidance law and not on the scenario under analysis. In this particular case, a pericenter raising is typically planned using a simple law that maximizes the time-rate of the pericenter radius. Nevertheless, the characteristics of the platform and the addition of a constraint on the apocenter radius transform such a classical approach into an over-complicated solution.

 




\section{Material and Methods}

\emph{Reinforcement Learning} \cite{kaelbling1996rl} is an area of \emph{Machine Learning} focused on studying optimization problems where an entity (agent) iteratively acts on a dynamical system (environment) to maximize a particular metric (rewards), with the goal to learn an optimal set of actions (policy). Contrary to \emph{Supervised Learning}, a \emph{Reinforcement Learning} algorithm does not need pre-generated data. Instead, it learns interacting directly with the environment.

\begin{figure}[!t]
\centering
\includegraphics[width=2.5in]{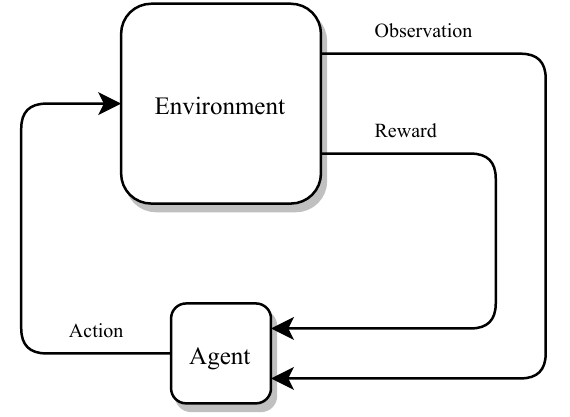}
\caption{\emph{Reinforcement Learning} model.}
\label{figure:rl}
\end{figure}

The communication between the agent and the environment is performed as shown in Figure \ref{figure:rl}: the environment feeds the initial observation to the agent. In each step, the agent decides the best action to take and informs the environment about it. Then the environment rewards (or penalizes) the latest action and returns the new observation to the agent. This process is repeated until the environment considers that the scenario is over. All the steps together are considered an episode.

Through exploring different actions in different episodes, the agent is able to identify which actions produce better rewards, effectively learning a control logic. As in every trial-and-error method, there needs to be a balance between exploration and explotation: an agent that explores too much does not typically focus on fine-tuning any particular solution in the search for the local minimum, whereas an agent too aggressive could miss a better solution in an unexplored area.

\subsection{Environment Definition}

The scenario under study is a constrained pericenter\footnote{Pericenter is the closest point of the orbit from the Earth} raising of a medium-Earth-orbit (MEO) satellite with an orbital plane tilted with respect to the principal axis of the inertial \emph{Geocentric Celestial Reference Frame} (GCRF). The satellite has an inertial orientation with three pairs of thrusters aligned with the principal directions of GCRF\footnote{Traditionally, satellites have one main engine for orbit control that is oriented towards the selected direction. Nevertheless, since the orientation is set by the needs of the maneuver, it prevents any other activity from happening at the same time. In order to prevent that off-time during the long maneuvering periods, the satellite under study has a configuration with three pair of thrusters where no special orientation of the satellite is needed.}. The full set of forces and other environment parameters can be found in Table \ref{table:environment}. In order to prevent the agent from learning a guidance law purely depending on the time, the initial mean anomaly of the satellite is selected randomly at the beginning of each episode.

\begin{figure}[!t]
\centering
\includegraphics[width=2.5in]{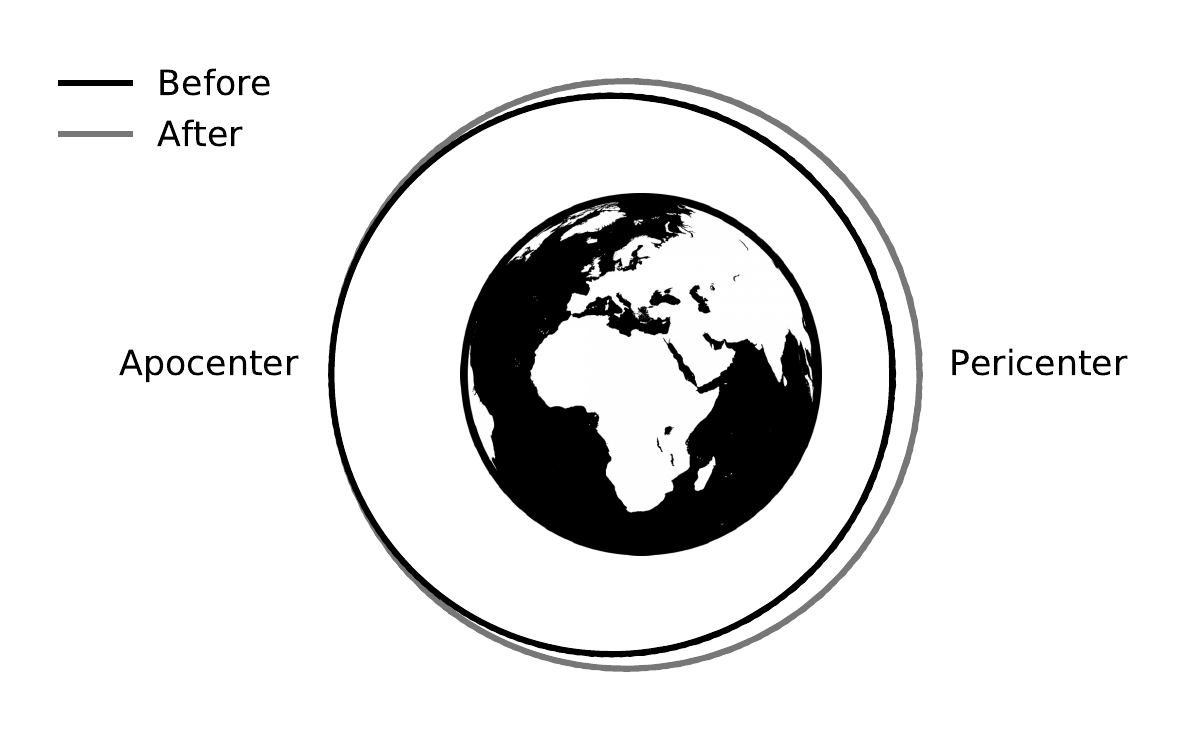}
\caption{Pericenter raising scenario of a MEO satellite.}
\label{figure:orbit}
\end{figure}

Each temporal step has the same duration. During that step, the force of each of the thrusters is constant and selected by the agent. Each episode has a fixed number of steps\footnote{A fixed number of steps is needed to obtain an unbiased measure of the achieved goal at the end of the episode. If the episode length was variable, longer episodes would get better rewards for the same control law.}, long enough to proof the agent across a whole number of revolutions of the orbit\footnote{\label{footnote:whole_orbits}The episode duration has been chosen to approximately represent a whole number of revolutions to reduce the dependence of the reward due to a different initial state.}, but short enough not to drag the episode unnecessarily. Rewards are given only at the end of the episode depending on the final satellite state. The reward has three components. First, reward is higher the more the pericenter is raised. Second, there is a penalty for mass consumption to prevent the agent from over-consuming fuel, e.g. firing the thrusters in directions irrelevant to the main objective. Finally, the environment penalizes any change in the apocenter\footnote{Apocenter is the furthest point of the orbit from the Earth.} radius. This penalty effectively enforces a soft constraint in the problem\footnote{This is a soft constraint because the agent could potentially learn to violate it if the overall reward is greater than the penalty due to the constraint. In practice, the optimum solution to the problem under study does not compromise the constraint that way.}.

\begin{table*}[!t]
\renewcommand{\arraystretch}{1.3}
\caption{Parameters Used in the Environment}
\label{table:environment}
\centering
\begin{tabular}{clrl}
\cline{2-4}
& Magnitude & Value & Units \\
\hline
\multirow{4}{*}{\shortstack{Dynamical parameters \\ for training}}
& Earth gravity field & Point mass & \\
& Third bodies & None & \\
& Solar radiation pressure & None & \\
& Thruster error & None & \\
\hline
\multirow{4}{*}{\shortstack{Dynamical parameters \\ for evaluation}}
& Earth gravity field & Spherical, $16$x$16$ & \\
& Third bodies & Sun and Moon & \\
& Solar radiation pressure & Yes & \\
& Thruster error & N(0, 10) & $\lbrack \% \rbrack$ \\
\hline
\multirow{4}{*}{\shortstack{Numerical \\ integration}}
& Method & Dormand-Prince & \\
& Order & $8$ ($5$, $3$) & \\
& Step size & $\lbrack 1, 1000\rbrack$ & $\lbrack s\rbrack$ (adaptative)\\
& Maximum position error & $1$ & $\lbrack m\rbrack$ \\
\hline
\multirow{8}{*}{\shortstack{Initial \\ state vector}}
& $t_0$ & 2022-06-16 00:00:00 & $\lbrack$UTC$\rbrack$ \\
& $r_{a_0}$ & $11000$ & $\lbrack km\rbrack$ \\
& $r_{p_0}$ & $9000$ & $\lbrack km\rbrack$ \\
& $i_0$ & $\pi/3$ & $\lbrack rad\rbrack$ \\
& $\Omega_0$ & $2\pi/3$ & $\lbrack rad\rbrack$ \\
& $\omega_0$ & $4\pi/3$ & $\lbrack rad\rbrack$ \\
& $M_0$ & $U(0 , 2\pi)$ & $\lbrack rad\rbrack$ \\
& $m_0$ & $100$ & $\lbrack kg\rbrack$ \\
\hline
\multirow{4}{*}{\shortstack{Satellite \\ parameters}}
& $F_{max}$ & $10$ & $\lbrack mN\rbrack$ \\
& $I_s$ & $4000$ & $\lbrack s\rbrack$ \\
& Area & $1$ & $\lbrack m^2\rbrack$ \\
& Reflection coef. & 2 & $\lbrack -\rbrack$ \\
\hline
\multirow{5}{*}{\shortstack{Scenario \\ parameters}}
& $\Delta t$ & $5$ & $\lbrack min\rbrack$ \\
& $N$ & $166$ & $\lbrack$steps$\rbrack$ $(\sim 5$ orbits$)$\\
& $W_{m}$ & $0.1$ & $\lbrack km^{-1}\rbrack $ \\
& $W_{r_a}$ & $0.1$ & $\lbrack km^{-1}\rbrack $ \\
& $W_{r_p}$ & $20$ & $\lbrack kg^{-1}\rbrack $ \\
\hline
\end{tabular}
\end{table*}

The main goal of this study is to analyse the behaviour of a dynamics-agnostic agent acting in that environment. In order to achieve that, one of the main points to avoid is for the environment to leak any kind of dynamic-related information to the agent, either explicit in the data themselves or implicit in the data representation. The interaction between the environment and the agent is performed through three channels: observations, actions and rewards.

Each observation contains the full state vector of the satellite: eight magnitudes representing time, position ($3$), velocity ($3$) and mass. Time (or rather elapsed time) was included following the results of \cite{pardo2018time} to prevent the violation of the Markov property. The position and velocity are expressed in GCRF in Cartesian coordinates. Using any other type of coordinates traditionally used in orbital dynamics, such as Keplerian elements or spherical coordinates, would give the agent some unintended knowledge of the dynamics of the problem. Finally, observations are normalised in the range $[-1, 1]$ using characteristic magnitudes of the problem, that is, total episode duration, maximum expected distance, maximum expected velocity, and initial mass.

The action consists of three magnitudes representing the normalised force (in the interval $[-1, 1]$) performed by each of the three pairs of thrusters of the satellite. As mentioned before, the orientation of the satellite is assumed constant in an inertial frame, and each pair of thrusters is installed parallel to each of the main directions of GCRF. This frame is not aligned with any relevant direction for the problem dynamics nor the orbit of the satellite.

The reward is one for the whole duration of each episode. The single reward is provided after the last action of the episode is performed. No intermediate or partial rewards are given to prevent biasing the learning of the agent towards those checkpoints. Note that, when computing the reward, there is no \emph{ground truth} to compare against; instead, the reward is given depending purely on the fulfillment of the objectives, the larger the better those objectives are achieved. The reward is computed as shown in Equation \ref{equation:reward}. Such a function rewards any increment of the pericenter radius, while penalises any change in the apocenter radius and any mass consumption. The weights $W_i$ for each component can be seen in Table \ref{table:environment} and have been selected to describe a trade off between pericenter raising and mass consumption, while keeping the reward value in a unitary order of magnitude.

\begin{equation}
\label{equation:reward}
R_i = 
\begin{cases}
0, & i \neq N \\
W_{r_p}\Delta r_{p} - W_{r_a} \abs{\Delta r_{a}} + W_m \Delta m, & i = N \\
\end{cases}
\end{equation}

For the purpose of this study, a simulator has been developed to interact with the agent. The simulator consists of an orbit propagator that integrates numerically the trajectory that the satellite would have had if it performed the selected actions. The simulator has been developed in Python using the Java-based orbital mechanics library \emph{Orekit} \cite{maisonobe2010orekit}. The environment has been chosen to follow \emph{OpenAI}'s \emph{Gym} specification \cite{brockman2016openai}, which allows the use of standard libraries and eases the possibility for other researchers to reuse it. The environment's source code and installation instructions have been published in \emph{GitHub}\footnote{\url{https://github.com/zampanteymedio/gym-satellite-trajectory}}.

\subsection{Agent}

The election of the agent is of the utmost importance for the resolution of the problem. This study assumes that the agent does not have any prior knowledge, implicit or explicit, of the dynamics of the problem or the shape of the reward. Therefore, the chosen actor must follow a model-free pattern, also called trial-and-error algorithms.

In the last few years, there has been a great development of different model-free algorithms to improve both the obtained solution and the time necessary to reach it. Those algorithms are divided in two groups: \emph{policy-based algorithms} seek to directly obtain and refine an optimum guidance law, whereas \emph{value-based} algorithms go after an understanding of the future rewards (or accumulated reward) at each possible observation after each possible action has been taken.

The selected algorithm is the \emph{Advantage Actor Critic} (A2C), a synchronous variant of the \emph{Asynchronous Advantage Actor Critic} (A3C) algorithm \cite{mnih2016asynchronous}. It leverages the benefits of both policy-based and value-based algorithms by estimating two functions on each step: one for the policy (actor) and one for the value (critic). On top of that, it supports continuous observation spaces and continuous actions, necessary for acting in the environment under study.

\begin{figure*}[!t]
\centering
\subfloat[]{\includegraphics[width=2.5in]{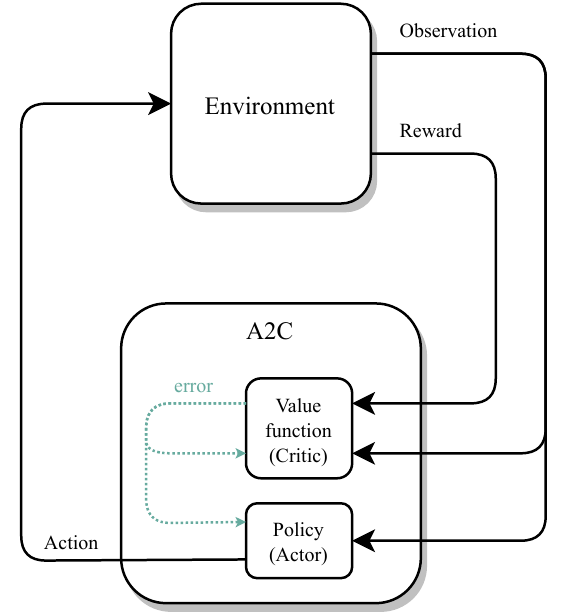}%
\label{figure:a2cenv}}
\hfil
\subfloat[]{\includegraphics[width=2.5in]{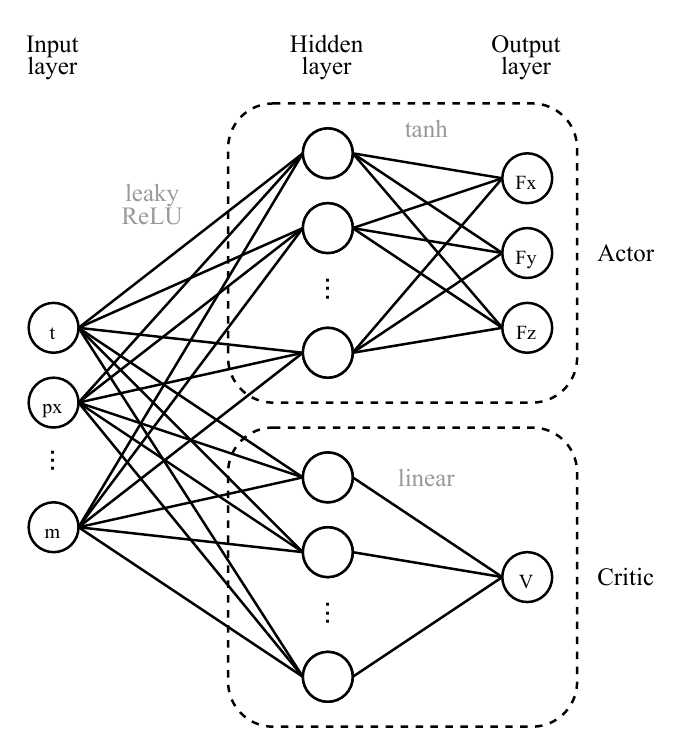}%
\label{figure:a2cnn}}
\caption{A2C agent.}
\label{figure:a2c}
\end{figure*}




For the purposes of this study, both actor and critic estimators have been modeled as fully-connected neural networks with 8 nodes in the input layer (observations) and one hidden layer with 200 nodes, as represented in Figure \ref{figure:a2cnn}. As showed in \cite{hornik1991approximation}, this configuration of feed-forward neural network is a universal approximator that can learn continuous mappings over compact input sets.

The activation function for the hidden layer is \emph{leaky ReLU}. It introduces non-linearities in the network and it prevents zones with zero slope, which could kill the training process. For the actor estimator, the output layer has three nodes (one per action dimension) with a \emph{tanh} activation function. This choice normalizes the actions in the interval $[-1, 1]$, as the environment requires. For the critic estimator, the output layer has one node with a linear activation function to keep it unbounded. Although \emph{A2C} allows for some of the hidden layers to be shared between the actor and the critic estimators, that capability was not considered necessary in this study due to the simple nature of the problem under study.

Due to the particular reward pattern of the environment, the discount factor of the agent has been set to $1$. This value prevents the reward obtained in the last episode from fading when accounting for it in the early steps of the episode. The learning rate has been adjusted after a manual exploration to be large enough to reduce the computation time and small enough to be able to refine the solution around the found minimum. The chosen algorithm to optimise the parameters of the neural networks is the Adam optimiser, due to its good general behaviour for this kind of problems. Other relevant parameters for the A2C agent are specified in Table \ref{table:a2c}.

The \emph{A2C} algorithm has been implemented using \emph{PyTorch} and the open-source library stable-baselines3 \cite{stable-baselines3}. As a proven solution widely chosen by the research community, using it reduces the probability of bugs and boosts the development speed. That allows the researcher to focus on the problem to solve and not on the algorithm used to solve it. All the code used for this study is published in \emph{GitHub}\footnote{\url{https://github.com/zampanteymedio/phd-satellite-trajectory-public}}.

\section{Results}

\subsection{Training Process}

Table \ref{table:a2c} shows the selected agent parameters after the parameter tuning exercise. On the one hand, some parameters have been pre-selected based on the unique characteristics of the problem. The discount factor $\gamma$ has been set to $1$ so that the steps at the beginning of the episode do not lose the information about the reward provided at the end. On the other hand, other parameters have been set after performing a manual exploration. The number of nodes in the hidden layer has been set to 200 as a compromise between computation speed, training speed and quality of the solution. Higher number of nodes only provides a marginal gain, whereas a lower number of nodes degrades the quality of the solution and requires a higher number of steps to reach the solution\footnote{An agent with a hidden layer of only 10 nodes eventually learns a fairly good behaviour after more than 2 million steps.}. The learning rate $\alpha$ has been set to $10^{-3}$, since higher rates lead to instability and lower rates just slow down the convergence.

\begin{table}[!t]
\renewcommand{\arraystretch}{1.3}
\caption{Parameters Used in the Environment}
\label{table:a2c}
\centering
\begin{tabular}{cll}
\cline{2-3}
& Parameter & Value \\
\hline
\multirow{3}{*}{A2C parameters}
& $\gamma$ & $1.0$ \\
& $\alpha$ & $10^{-3}$ \\
& optimizer & Adam \\
\hline
\multirow{3}{*}{Actor network}
& nodes in hidden layer & $200$ \\
& act. fun. in hidden layer & leaky ReLU \\
& act. fun. in output & tanh \\
& distribution for output & diagonal Gaussian \\
\hline
\multirow{3}{*}{Critic network}
& nodes in hidden layer & $200$ \\
& act. fun. in hidden layer & leaky ReLU \\
& act. fun. in output & linear \\
\hline
\multirow{1}{*}{Learning parameters}
& number of training steps & $5 \cdot 10^5$ \\
\hline
\end{tabular}
\end{table}

\begin{figure}[!t]
\centering
\includegraphics[width=2.5in]{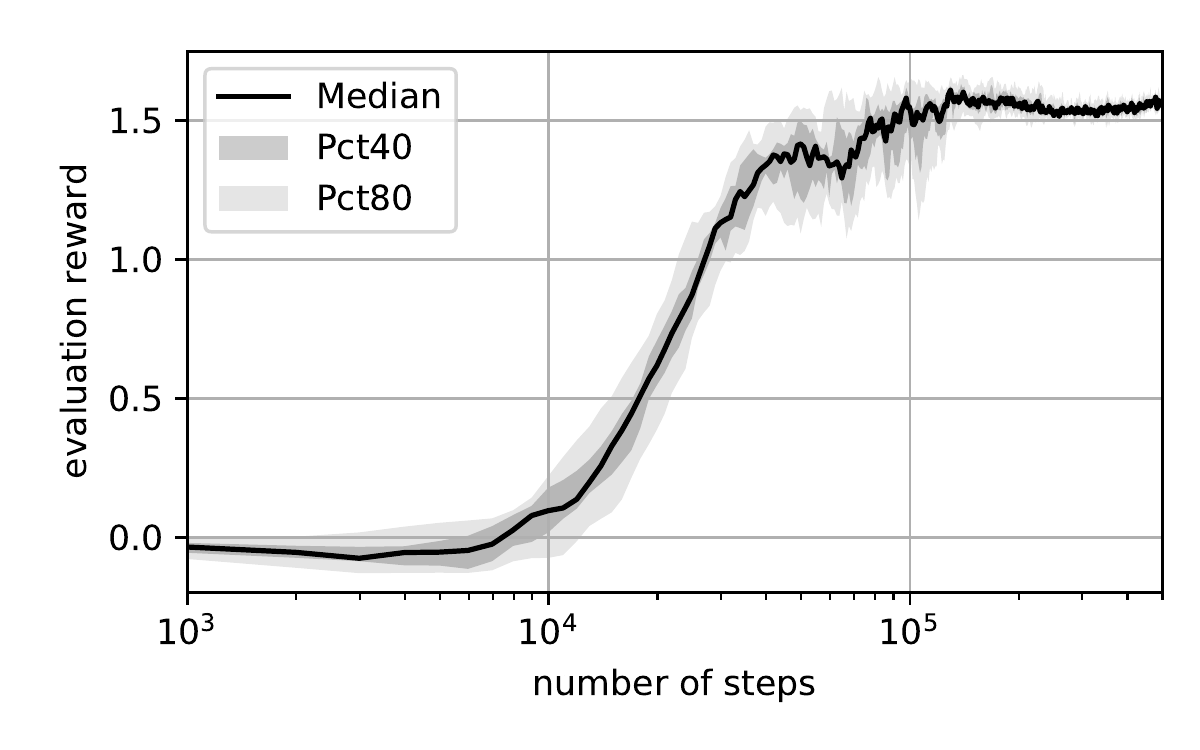}
\caption{Evaluation during training for 14 agents.}
\label{figure:training}
\end{figure}

In order to reduce the uncertainty due to random behaviours typically associated with AI models, 14 agents with different random seeds have been trained. The rewards obtained by the evaluation of the agents mid-training are shown in Figure \ref{figure:training}\footnote{After every 5 training episodes, the agent is evaluated using a new episode and the reward is obtained and plotted. This evaluation episode is not used for training purposes.}. There is a first phase before $10^5$ steps where agents are in an exploration phase and are able to learn how to increase the reward. Beyond $10^5$ steps, agents stop learning and only replay solutions very close to the optimal policy. After $5 \cdot 10^5$ steps, all the agents arrive to a similar solution very close to the best ever achieved\footnote{The best reward ever achieved by any agent is $1.68$. This corresponds to an increase of the pericenter radius of $24.0 km$ using $35 g$ of fuel and with a deviation of $0.2km$ of the apocenter radius.}. Although increasing the number of steps could bring that final agent all the way up to the best ever achieved solution, the gain of doing so would be marginal and the computational cost excessive.

\subsection{Planned Trajectory}

\begin{figure*}[!t]
\centering
\subfloat[]{\includegraphics[width=2.5in]{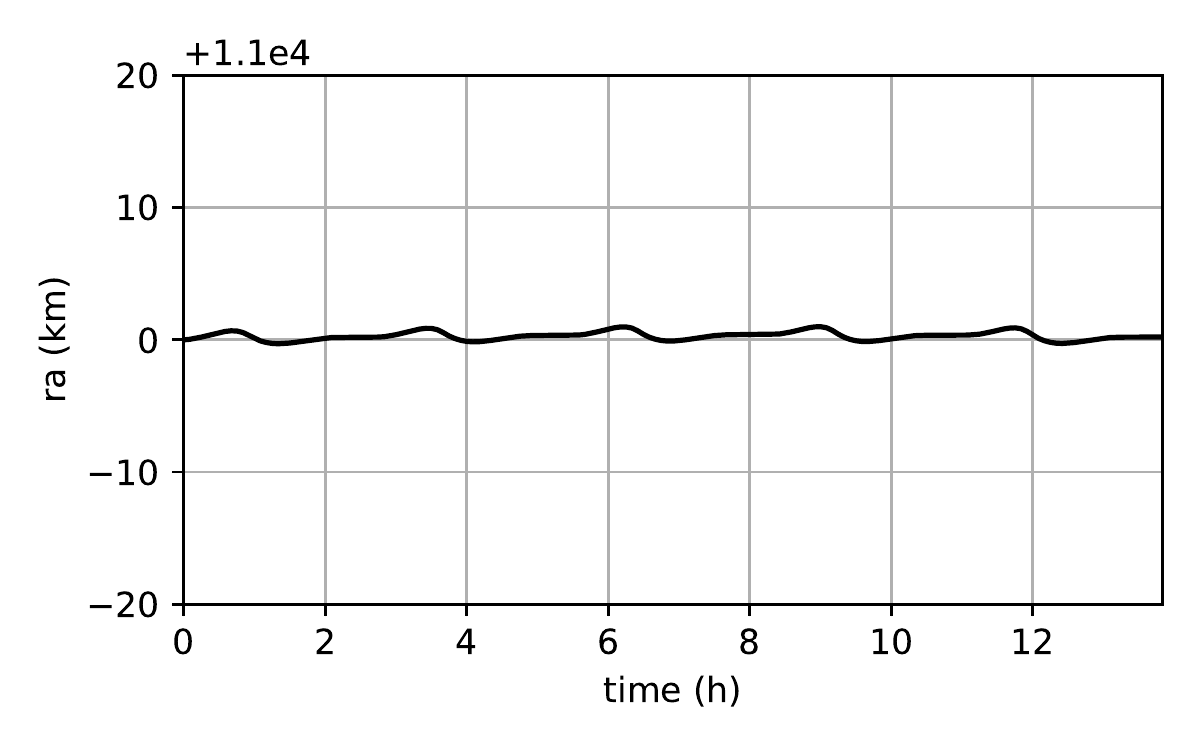}%
\label{figure:planned_ra}}
\hfil
\subfloat[]{\includegraphics[width=2.5in]{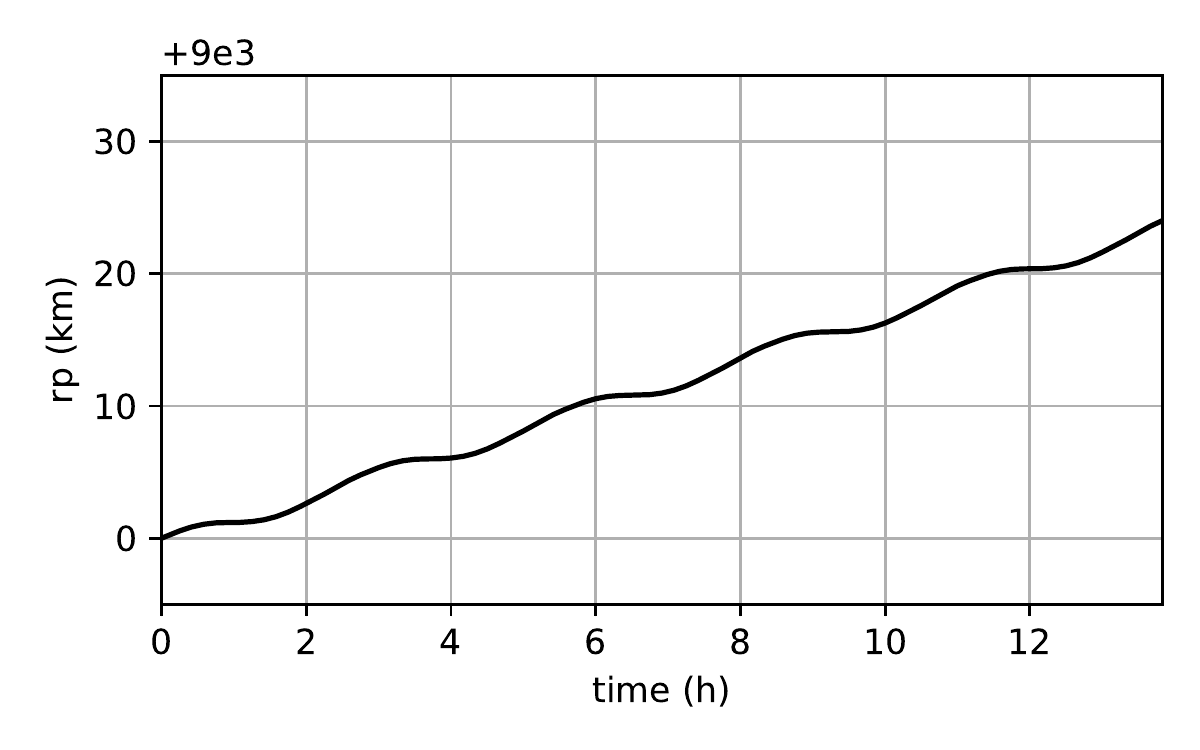}%
\label{figure:planned_rp}}
\hfil
\subfloat[]{\includegraphics[width=2.5in]{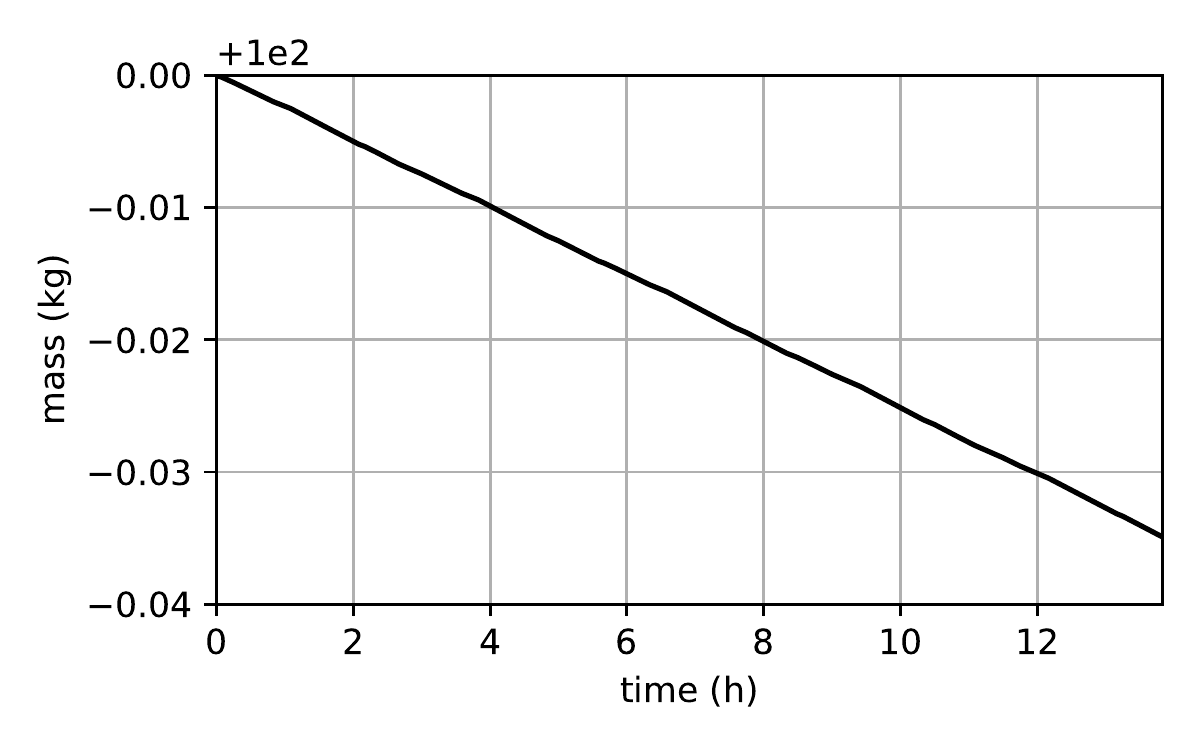}%
\label{figure:planned_mass}}
\hfil
\subfloat[]{\includegraphics[width=2.5in]{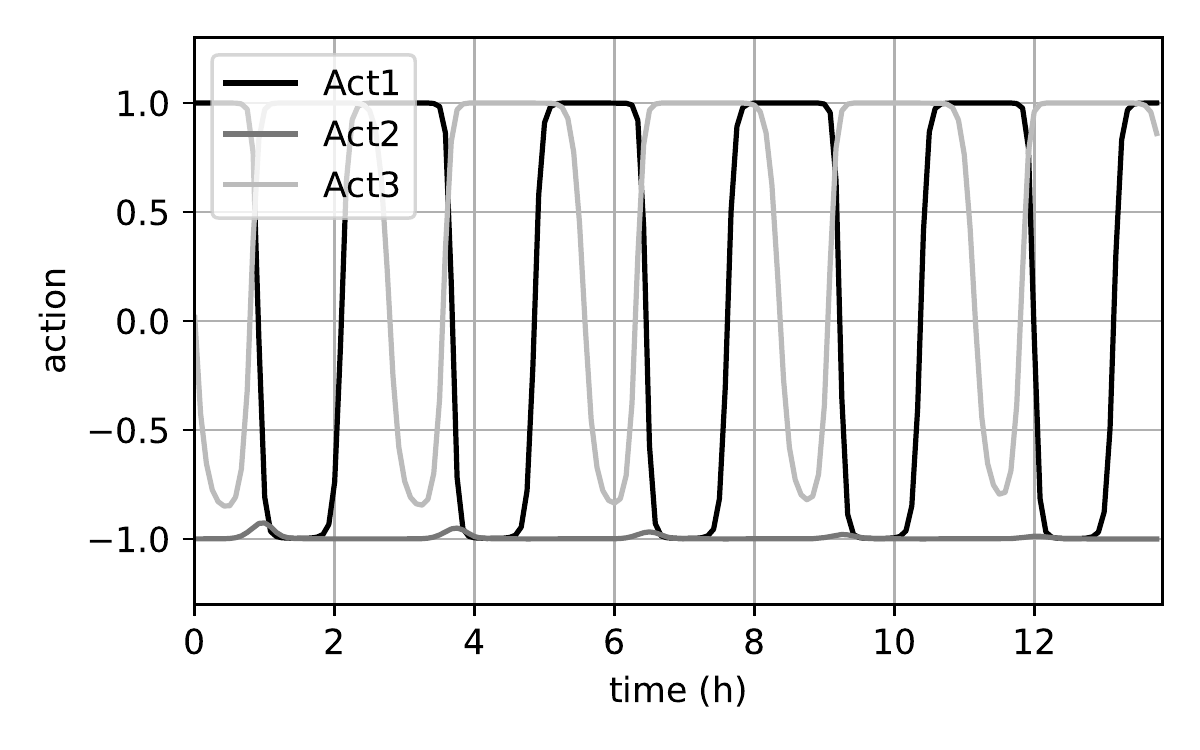}%
\label{figure:planned_action}}
\caption{Trajectory planned by the selected agent.}
\label{figure:planned}
\end{figure*}

After the training exercise, the best solution ever achieved by any agent has been selected and used to generate a trajectory in an environment without perturbations. As Figure \ref{figure:planned} shows, the apocenter radius stays controlled around the initial value while the pericenter radius monotonically increases. Given that thrusters are constantly firing in all directions, mass decreases over time.

Figure \ref{figure:planned_action} shows the actions taken by the agent along the episode. Since an episode consists of five full orbits\footnote{See footnote \ref{footnote:whole_orbits}.}, the agent applies five times a similar thrusting pattern. During each orbit, one of the thrusters is constantly firing nearly at full force in one direction, while the other two alternate magnitude and direction in different moments of the orbit. Although this solution naturally appears as the control commanded by the trained agent, it is very challenging to model this kind of patterns with a full analytic or parametric law, bringing out the power of \emph{Reinforcement Learning} compared to traditional methods.

\subsection{Performance during Cruise}

The final step on the evaluation of the agent is to assess its performance if used as a satellite controller in real time, in an environment that contains small perturbations with respect to the one the agent trained in. The dynamical model of the new environment considers the perturbations described in Table \ref{table:environment}: a non-spherical Earth, solar radiation pressure, and Sun and Moon gravity forces. On top of that, the thrusters of the satellite have a random mis-performance.

\begin{figure*}[!t]
\centering
\subfloat[]{\includegraphics[width=2.5in]{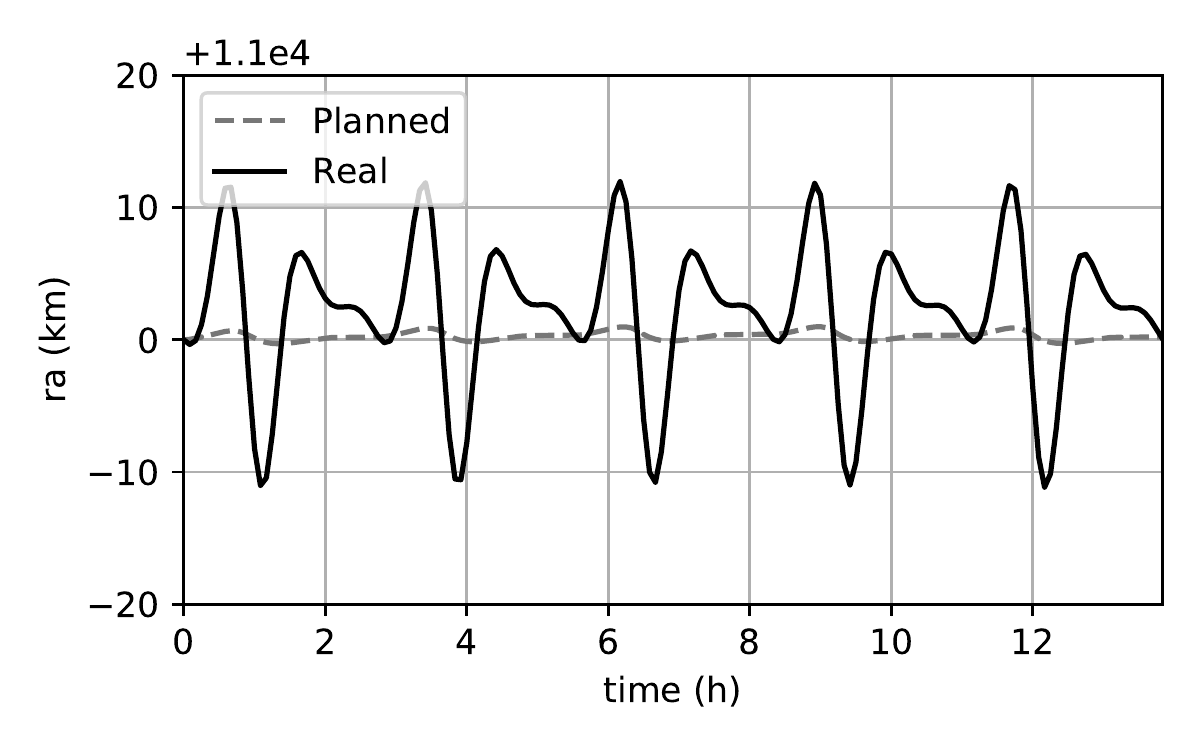}%
\label{figure:real_ra}}
\hfil
\subfloat[]{\includegraphics[width=2.5in]{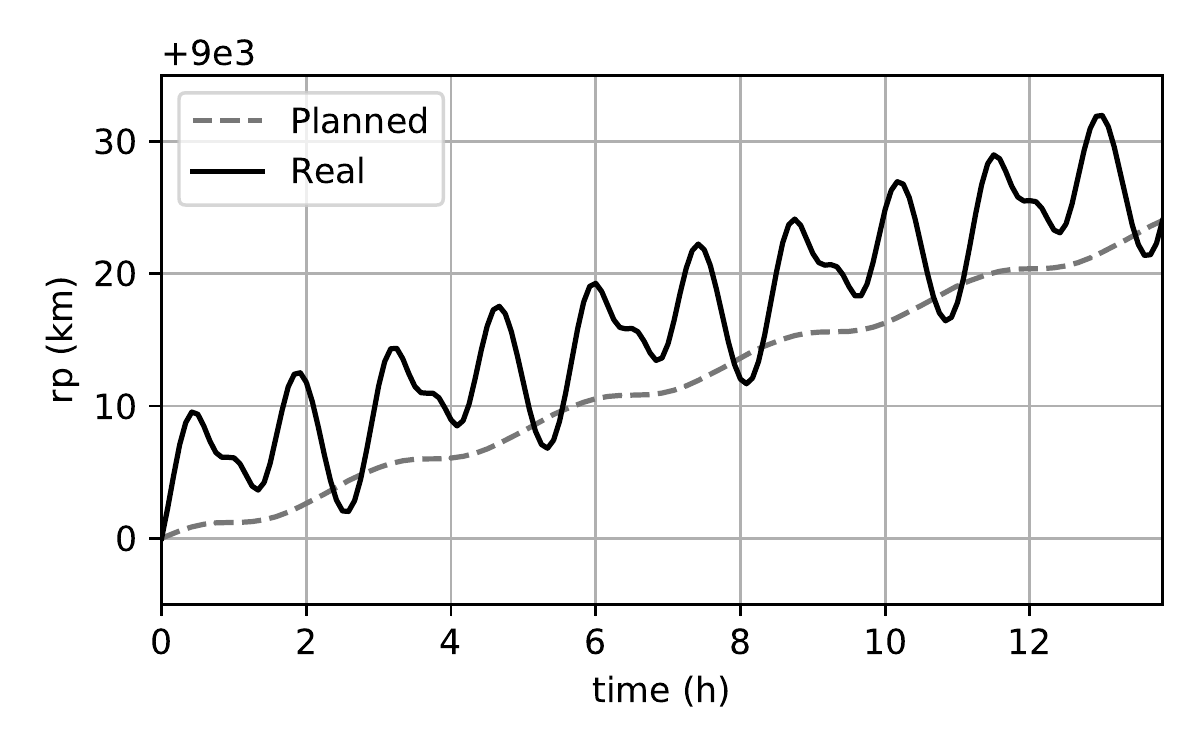}%
\label{figure:real_rp}}
\hfil
\subfloat[]{\includegraphics[width=2.5in]{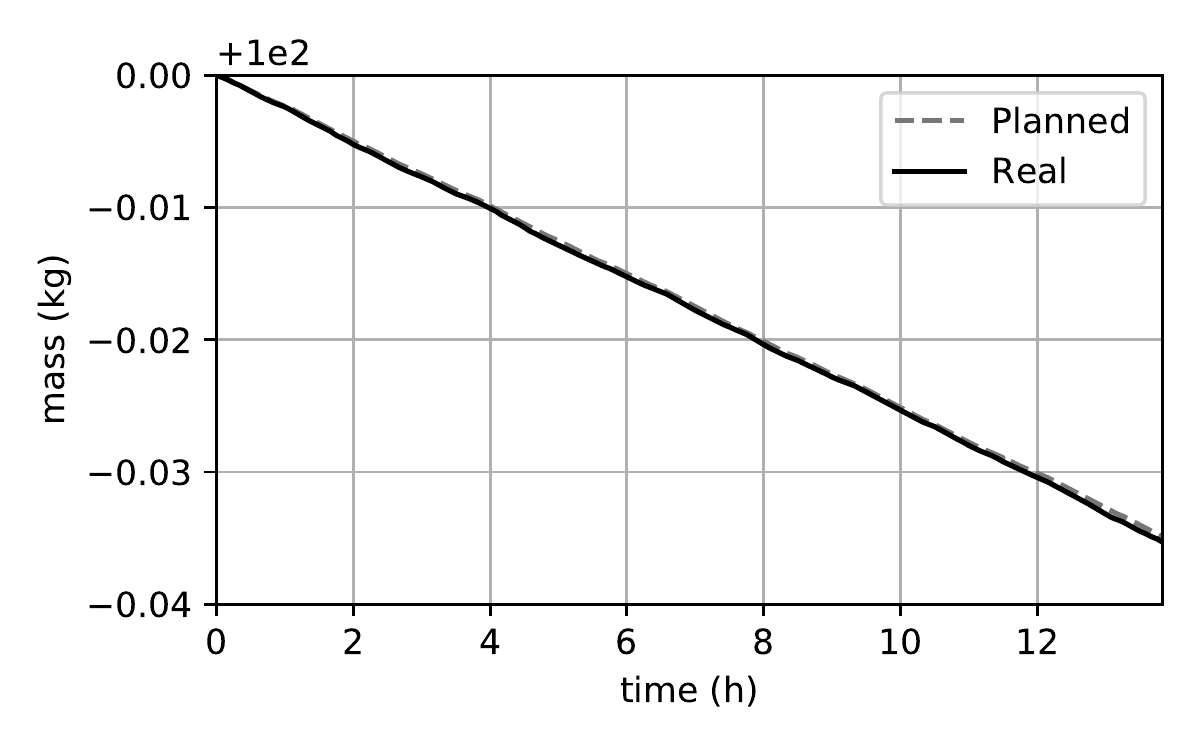}%
\label{figure:real_mass}}
\hfil
\subfloat[]{\includegraphics[width=2.5in]{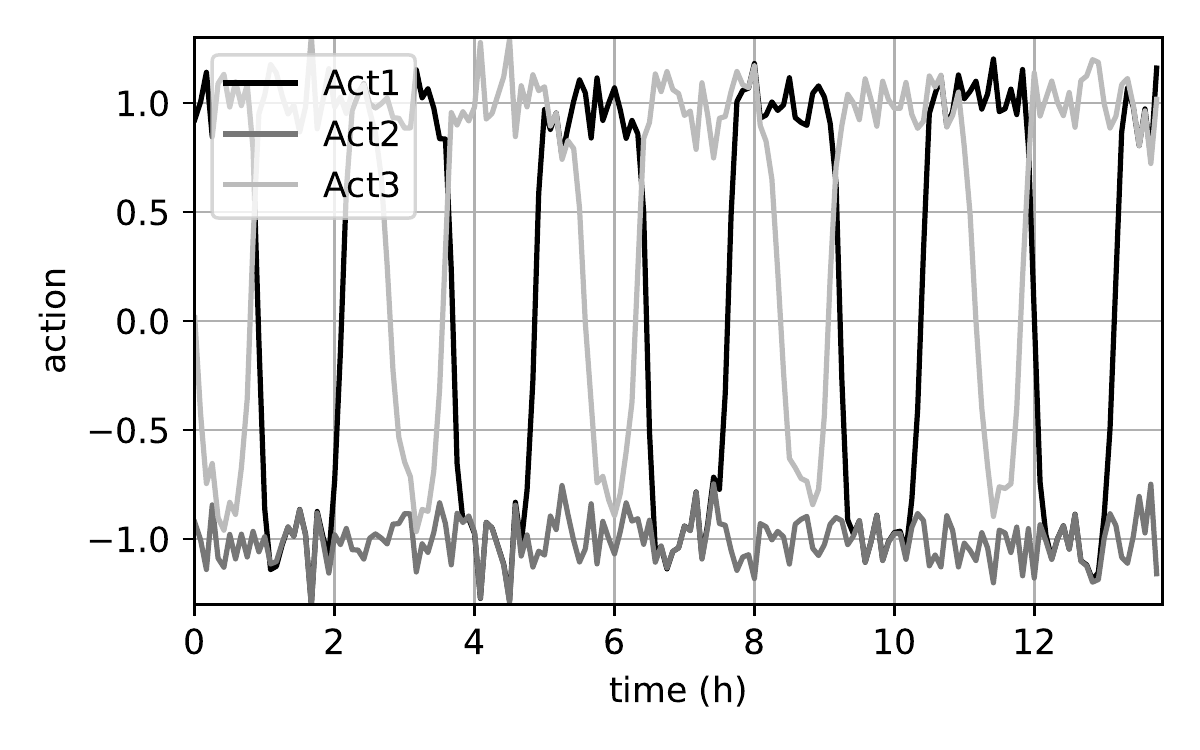}%
\label{figure:real_action}}
\caption{Trajectory achieved by the selected agent on an environment with perturbations.}
\label{figure:real}
\end{figure*}

Figure \ref{figure:real} shows the variation of the relevant parameters of the trajectory of the satellite. When plotting the osculating pericenter and apocenter radii\footnote{Due to the new forces in the environment, the orbital elements representing the instant (osculating) pericenter and apocenter radii are not constant even in absence of thrust. Instead, they periodically change along the orbit.}, it can be seen that the trend of those elements remains the same: apocenter radius remains stable while pericenter radius increases. Moreover, the increase rate is in line with the planned figure. Therefore, the agent is performing as planned in an environment even if it was not exactly the same environment used for training, opening the possibility to be used as a real-time on-board controller. This also shows that simplified environments can be considered during training as long as they capture the main characteristics of the real environment, reducing the total computation time.

\section{Discussion}

This study demonstrates that \emph{Reinforcement Learning} is a valid tool for the optimisation of the trajectory of satellites equipped with a low-thrust propulsion system. It also proves that an \emph{a priori} knowledge of the dynamics of the problem or an intuition of the optimum are not necessary for an agent to learn an optimal control law.

The guidance law produced by a trained agent appears naturally as a result of its training. Nevertheless, it would have been very challenging to model this kind of guidance with a full analytic or parametric law, bringing out the power of \emph{Reinforcement Learning} compared to traditional methods.

The simplicity of the approach described in this study is in clear opposition to all the studies performed so far. Most of the other methods using traditional optimization or \emph{Artificial Intelligence} share the need to have a strong knowledge of the problem, its optimal solution or a ground truth, such as a previously-generated optimal set of actions or a reference trajectory. This defeats the purpose of using \emph{Artificial Intelligence} because the agent will just replicate that ground truth without challenging its purpose and without trying to find better ways to achieve the wanted result. Moreover, that blind faith of the agent in the ground truth could lead to unexpected situations like fuel over-consumption due to the desire to follow a non-physically-feasible reference trajectory.

This study focuses on solving the problem in question using the simple original needs: increase of the pericenter radius and minimizing the fuel consumption, while keeping the apocenter radius constant. No previous data are required to solve the constrained optimization problem since the agent generates its own data from the environment during training and it autonomously learns what the best approach to achieve the original needs is. During the control phase, the agent does not blindly follow the planned trajectory; instead, it adapts the trajectory keeping the original needs in mind.

In order to create, train and use the agent described in this study, there is no need to have a deep understanding in maths or space dynamics. On top of that, the analysis has been performed using open-source tools and no expensive computational resources. This effectively broadens the profile of researchers and engineers that can work on the field of satellite trajectory optimization and reduces the costs associated to satellite control.

On the other hand, totally ignoring the knowledge on the dynamics of the problem is a too radical approach. This knowledge could potentially be used by future studies or operational systems to shape the agent and make it learn faster, better and in a more general set of scenarios. In the case under study, the use of Keplerian elements as input parameters would seem like the obvious choice, since they contain most of the information about the dynamics of the problem. Nevertheless, this kind of decisions needs to be assessed with care: a bias preventing the agent from exploring freely could be introduced, resulting in a sub-optimal solution.

Finally, this study proves that the same agent used to obtain the planned trajectory can be used to autonomously control the propulsion system in real-time, further reducing the operational costs of the satellite. Although preliminary results are very positive, this approach still needs to be tested thoroughly before it can be used operationally due to the high value at risk inherent to the space sector: overusing on-board resources or even losing the satellite.

\section{Conclusion}

This study demonstrates that \emph{Reinforcement Learning} is a promising method for satellite trajectory optimization. It has been shown that it is not strictly necessary to know the dynamics of the problem or an \emph{a priori} solution in order for a trained agent to learn a quasi-optimal thrusting law. It has also been presented how such a trained agent can be used to both plan a trajectory and to control the satellite along that trajectory in real time, even when unforeseen perturbations occur. This has been achieved using an agent with a surprisingly low number of parameters, well far from the millions of parameters used in other areas such as artificial vision or speech recognition, which leads to think that a similar approach could be used for more complex scenarios without a great increment of the needed computational power.

There are a few aspects of the study that could lead to better solutions or faster training times before the proposed approach is used operationally. For example, a more exhaustive hyper-parameter tuning campaign, other neural network configurations or alternative \emph{Reinforcement Learning} algorithms could be analysed.

Overall, this study opens the door to using \emph{Reinforcement Learning} to more complicated problems where it is very challenging to describe an optimal solution analytically, such as interplanetary flight arcs, or environments with non-Keplerian dynamics like Lagrangian points or around a comet. This approach could also be used to explore new strategies in other areas where solutions have room for improvement, such as collision avoidance, geostationary satellite co-location or multi-satellite formation.

\appendix[Nomenclature]

\begin{tabular}{ll}
$A_i$ & action taken by the agent at step $i$ \\
$F_{max}$ & maximum force of each thruster \\
$i$ & inclination \\
$I_s$ & specific impulse of the thrusters \\
$m_i$ & mass of the satellite at step $i$\\
$M$ & mean anomaly \\
$N$ & number of steps per episode \\
$r_a$ & apocenter radius \\
$R_i$ & \makecell[l]{reward given by the environment after the \\ ~~~~action at step $i$} \\
$r_p$ & pericenter radius \\
$S_i$ & \makecell[l]{observation representing the state of the \\ ~~~~environment at step $i$ } \\
$W_i$ & weight in the reward function for feature $i$ \\
$\alpha$ & learning rate of the agent \\
$\Delta t$ & duration of environment simulation step\\
$\gamma$ & discount factor used by the agent \\
$\Omega$ & right ascension of ascending node \\
$\omega$ & argument of pericenter
\end{tabular}


%





\ifCLASSOPTIONcaptionsoff
  \newpage
\fi



\bibliographystyle{IEEEtran}
%
\bibliography{main}
\end{document}